\relax
\documentclass[letterpaper]{article} %DO NOT CHANGE THIS
\usepackage{aaai19}  %Required
\usepackage{times}  %Required
\usepackage{helvet}  %Required
\usepackage{courier}  %Required
\usepackage{graphicx}  %Required

\usepackage{hyperref}

\usepackage{latexsym}
\usepackage{array}
\usepackage{color}

\usepackage{graphicx}
\usepackage[font=normalsize,position=bottom]{subfig}
\usepackage{algorithm}
\usepackage[noend]{algpseudocode}
\usepackage{multirow}

\usepackage{amssymb}
\usepackage{bm}
\usepackage{balance}
\usepackage{arydshln}
\usepackage{mathtools}
\usepackage{soul}

\usepackage{booktabs}
\usepackage{amsmath,amsfonts}
\usepackage{graphicx}
\usepackage{xspace,color}

\usepackage{bbm}

\usepackage{xspace}
\usepackage{textcomp}

\usepackage{tabularx}
\newcolumntype{L}{>{\raggedright\arraybackslash}X}

\newcommand{\eg}{\emph{e.g.,}\xspace}
\newcommand{\ie}{\emph{i.e.,}\xspace}
\newcommand{\specialcell}[2][l]{%
  \begin{tabular}[#1]{@{}l@{}}#2\end{tabular}}

\begin{document}
\title{Generative Stock Question Answering}

\author{
Zhaopeng Tu\thanks{Zhaopeng Tu is the corresponding author. Work was done when Yong Jiang and Lei Shu were interning at Tencent AI Lab.}\\
Tencent AI Lab\\
zptu@tencent.com
\And
Yong Jiang\\
ShanghaiTech University\\
jiangyong@shanghaitech.edu.cn
\AND
Xiaojiang Liu\\
Tencent AI Lab\\
kieranliu@tencent.com
\And
Lei Shu\\
University of Illinois at Chicago\\
lshu3@uic.edu
\And
Shuming Shi\\
Tencent AI Lab\\
shumingshi@tencent.com
}

\maketitle
\begin{abstract}

We study the problem of stock related question answering (StockQA): automatically generating answers to stock related questions, just like professional stock analysts providing action recommendations to stocks upon user's requests.
StockQA is quite different from previous QA tasks since (1) the answers in StockQA are natural language sentences (rather than entities or values) and due to the dynamic nature of StockQA, it is scarcely possible to get reasonable answers in an extractive way from the training data; and (2) StockQA requires properly % interactions between QA pair and 
analyzing the relationship between keywords in QA pair and the numerical features of a stock.
We propose to address the problem with a memory-augmented encoder-decoder architecture, and integrate different mechanisms of number understanding and generation, which is a critical component of StockQA.

We build a large-scale dataset containing over 180K StockQA instances,\footnote{The data is publicly available at \url{http://ai.tencent.com/ailab/nlp/data/stockQA.tar.gz}.} based on which various technique combinations are extensively studied and compared. Experimental results show that a hybrid word-character model with separate character components for number processing, achieves the best performance. 
By analyzing the results, we found that 44.8\% of answers generated by our best model still suffer from the {\em generic answer} problem, which can be alleviated by a straightforward hybrid retrieval-generation model.
% incorporating external retrieval answers with a straightforward multi-source model.

\end{abstract}

\section{Introduction}

%Enabling computers to answer questions is a long-standing research problem ever since 1960s.
Question answering (QA) has been a long-standing research problem in natural language processing (NLP) since 1960s.
%In this paper, we are interested in how to answer general stock questions given the dynamic stock knowledge base.
In this paper, we focus on a sub-type of QA tasks: answering stock related questions.
Our goal is to automatically generate convincing natural language answers to stock related questions, just like the answers given by the professional stock analysts.
Table~\ref{examples-qa} shows two representative StockQA examples. 
% Examples of stock questions include ``I bought CJGD at \$12, what should I do now?" and ``Morning, experts! I have 30\% position of LJJK at \$12.6, 37\% loss, should I cut loss now? Thanks!"
%We found this problem setting have the following features which make it quite different from the previous QA problems. 
Although fundamentally a QA problem, StockQA is quite different from those well studied in previous research due to the following two characteristics.

\begin{table}[t]
\centering
\renewcommand\arraystretch{1.1}
\begin{tabular}{m{0.45\textwidth}}
\hline
{\bf Q}: I bought TSLA at \$349, how to handle it?\\
\hdashline
{\bf A}: TSLA is in a weakening trend. I suggest you to sell some shares to reduce your risk. Sell all if it could not rise above the 50-day MA \$330.\\
\hline
{\bf Q}: What is the support level of MSFT? I invested 30\% of my money in it. \\
\hdashline
{\bf A}: MSFT is trying to break the previous high point. Hold it if it can stay steady on \$79.15. \\
\hline
\end{tabular}
\caption{Two representative StockQA examples.}
\label{examples-qa}
\end{table}

%Second, while most of the previous research focusing on answering questions using entities, either by selecting existing entities from questions or knowledge, nor by knowledge base inference. The stock questions we are studied in this paper are mostly why-questions. For examples, the questions are like "I have bought MSFT at price \$60, and should I sell them all?" "What do you think of APPL, will it rise in the next week?" Investors post these kinds of questions on forums to get advises from specialists or other investors. Different from the previously well-studied what-questions, why questions are hard to answer. Simple answering the question with "yes" or "sell it" will hardly convince the questioner. Most of the answers will not only include the conclusion or advises on actions, but also with some information and how they get the conclusion, such as "the stock price is currently high above the Moving Average 60, sell it" or "hold it until the it hits the pressure price \$78". 
First, StockQA is different from traditional QA tasks in obtaining answers.
In previous QA tasks, the answers are typically selected from a knowledge base (\eg entities)~\cite{berant2013semantic,Yin:2016:IJCAI} or question-answer pairs collected from web forums (\eg sentences)~\cite{bian2008finding,cong2008finding}. In contrast, due to the dynamic nature of stock related QA, we {\em generate} answers based on the question and the knowledge of the referred stock. In our task, the answers to a given question may vary according to the time when the questions are asked, as well as the stock referred to. 
Accordingly, the retrieval-based approaches cannot be applied in this task.

Second, number understanding and generation is critical in StockQA for generating reasonable answers.
In previous knowledge-base QA tasks~\cite{berant2013semantic,bordes2015large}, numerical values are either not involved or simply treated as attribute values of entities (just like other types of attribute values). For example, in answering questions about the elevation of a mountain, the numerical answer is generally selected from a knowledge base like other entities (\eg mountain names).
In StockQA, however, reasonable answers often contain numbers, which are generated by analyzing relations between numbers in question and stock numerical knowledge (\eg the numeric value of cost price in question and close price in knowledge base), as well as text words in question and numbers in stock knowledge (\eg the words ``resistance level'' in question and ``high price'' in knowledge base).
In addition, numbers in answers generally refer to an estimated price of support level or resistance level, which cannot be generated with an arithmetic operation (\eg subtraction or rounding) on numbers in the knowledge base, as done in~\cite{murakami2017learning}.

% are often the results of properly analyzing the relationship among these numerical values. 

% Examples of stock numerical knowledge include open price, close price, trading volume, and other stock indicators~\cite{murakami2017learning}.

In this work, we treat the StockQA task as a natural language generation problem, and address it with a memory-augmented encoder-decoder model. The encoder summarizes the question consisting of words and numbers, and the memory stores stock numerical knowledge including open price, close price, trading volume, and other stock indicators. The decoder generates answers by attending the representations from both the encoder and memory.
In response to the challenge of number processing, we propose different mechanisms for number composition and decomposition by exploiting their inside character information. Among them, a hybrid word-character model achieves the best performance by introducing two additional networks to consult character components whenever the model encounters a number.

\paragraph{Contributions.} We make the following contributions:
% \vspace{-5pt}
\begin{enumerate}
  \setlength{\itemsep}{4pt}
  \setlength{\parskip}{0pt}
  \setlength{\parsep}{0pt}
   \item We introduce a novel StockQA task and define the corresponding evaluation metrics. 
   % Compared to the QA tasks studied in the previous research, one key challenge of StockQA is understanding numbers in questions and generating numbers in answers.
   \item We collect a large-scale real-world dataset containing 180K instances, which will be released to offer public resources for this task.
   % based on which the performance of different technique combinations are extensively evaluated and compared. The dataset will be released, offering resources for training new QA models.
   \item We present an end-to-end framework for the task, which combines advanced sub-networks for sentence generation and number processing.
   % We employ a memory-augmented encoder-decoder model to generate answer text. One critical component is a hybrid word-character model which generates words and numbers separately to get more informative answers containing reasonable numbers.
   Experiments demonstrate the potential of the framework through both qualitative and quantitative evaluation. % of the generated answers.
   \item We reveal that benefiting from the informative retrieved answers, a simple hybrid retrieval-generation model can significantly improves the diversity and informativeness of the generated answers.
\end{enumerate}

\section{Problem Formulation and Dataset}

\paragraph{Problem Formulation.}
The challenge of StockQA lies in the dynamic nature of stock related QA, in which the QA pairs may vary according to the time when the questions are asked, as well as the stock referred to. %, as well as the enquirer. 
Hence, the system requires the ability to formulate different answers to the exact same question.
We model the dynamical property as a stock knowledge base, which describes the referred stock with numeric features being fetched at the time when the question is asked.
Specifically, the knowledge base $KB$ consists of a set of stock features, which are in the form of $<${\em feature}, {\em value}$>$ pairs with the values being float numbers, as shown in Table~\ref{table-stock-kb}.
The features could be prices, volumes, or some widely used indicators. 
In this paper, we do not consider the non-numeric features, such as the market news.
% industry of the company or the market news. 
We leave the exploration of such features as a future work.

\begin{table}[t]
\centering
\subfloat[Stock numerical features used in this work.]{
\begin{tabular}{c | l }
\textbf{Type} & \textbf{Description}\\
\hline
Price & \specialcell{Open, close, high, low, moving average over\\5 days (MovAvg5),MovAvg10, MovAvg20} \\
\hline	
Volume & \specialcell{Trading volume, average trading volume of \\5 days (AvgVol5), AvgVol10, AvgVol20}\\
\hline	
Change & \specialcell{Price change, change rate, turnover}\\
\end{tabular}
}\\
\subfloat[Example of some stock features.]{
\begin{tabular}{l | r || l | r}
\textbf{Feature} & \textbf{Value}  &  \textbf{Feature} & \textbf{Value}\\
\hline
Open  &   9.75    &   AvgVol5 &  53186.72 \\
Close &   9.15    &   AvgVol10 &  53186.72     \\
\cline{3-4}
High  &   9.93    &   Price change &  -0.45\\
Low   &   9.02    &   Change rate  &  -0.05\\
\end{tabular}
}
\caption{Stock numerical knowledge base.}
\label{table-stock-kb}
\end{table}

The goal of StockQA is to generate a natural language answer ${A}$ that could answer the question $Q$, given the corresponding stock knowledge base $KB$. 
Most of the questions in StockQA are ``{\em what}" and ``{\em how}" questions, such as asking for stock trend prediction (\ie what-type) and action recommendation (\ie how-type).
To generate reasonable answers, it is important to capture the relations between the indicators in question and the stock features in knowledge base, as well as the relative value of numbers. For example, to answer the question ``{\em what is the support level?}'', the model should first capture the pattern that ``support level'' generally corresponds to ``low price'' of a stock. Then, the generated answer is expected to contain a number, whose value is close to the low price of the stock stored in the knowledge base. As seen, the number to be generated cannot be calculated by an exact arithmetic operation (\ie subtraction), as done in ~\cite{murakami2017learning}. Instead, it requires an estimation operator to compare the value of numbers, and thus generate a number in a reasonable scale.

\begin{table}[t]
\centering
\begin{tabular}{c | l }
\textbf{Score} & \textbf{Heuristics}\\
\hline
0   &   poor Rel.\\
%0.25   &   normal Rel. $or$ low Pre. has factual error $or$ not fluent\\
0.25   &   normal Rel. $or$ low Flu. $or$ low Pre.\\
0.5   &   good Rel. $and$ poor Inf.\\
0.75   &   good Rel. $and$ normal Inf.\\
1.0   &   good Rel. $and$ good Inf.\\
\end{tabular}
\caption{Heuristics to combine the metrics. ``Rel'': relevance, ``Pre'': precision, ``Flu'': fluency, ``Inf'': informativeness.}
\label{metric}
\end{table}

\paragraph{Evaluation.}
Given that the generated answer is a natural language sentence instead of a single entity, we cannot measure its quality with precision and recall, which are commonly used in factoid QA tasks. 
As we aim at generating answers to convince the enquirers, we use two widely-used automatic metrics in language generation tasks: {\em diversity} and {\em informativeness}. 
% to measure if the answers are convincing. 
Diversity \cite{li-EtAl:2016:N16-11} measures the ratios of distinct n-grams of the answers, while  informativeness measures the number of unique n-grams of the answers.

In addition, we use four more manual evaluation metrics to measure the generated answer from four different perspectives: {\em relevance}, 
{\em fluency}, {\em precision}, and {\em informativeness}.
%\begin{itemize}
%    \item {\em relevance}: how relevant they are to the question;
%    \item {\em fluency}: whether they are fluent;
%    \item {\em precision}: whether the answers are against the fact; 
%    % Factual errors will make the answers wrong, and much less convincing;
%    \item {\em informativeness}: how informative the answers are.
%    %whether the answers are convincing; this is important when making argument to convince the questioners. For answers with some general reply or replies with only actions, the informativeness of the answer is poor.
%\end{itemize}
Among them, ``{\em precision}'' and ``{\em informativeness}'' are two specific metrics for StockQA, which set it apart from common knowledge base QA task. 
Precision measures whether the generated answer has factual errors. For example, if a user asks for an estimation of support level, while the answer offers a number with value being higher than the high price, we treat it as ``factual error''.
A reasonable and convincing answer generally contains sufficient information, including the interpretation of the recent trend, the prediction of future trend, and the action recommendations according to the user's request and the analysis. 
There are two ratings for {\em fluency} (\ie ``yes'' or ``no'') and {\em precision} (\ie ``has factual errors'' or ``no factual errors''), and three ratings (\ie ``poor'', ``normal'', or ``good'') for both {\em relevance} and {\em informativeness}.
Table~\ref{metric} shows the heuristic rules of mapping the metrics to a ranking score ranging from $0$ to $1$ with five uniform scales.
As seen, relevance is a zero-tolerance requirement for the generated answers. Above that, precision and fluency would serve as another threshold to filter out low-quality answers. Finally, informativeness is used as an indicator to identify high-quality answers. For example, generic answers, which are quite common in dialogue and QA tasks, are labelled as ``poor informativeness''.

\begin{table}[t]
\centering
\begin{tabular}{l | l }
\textbf{Property} & \textbf{Statistic}\\
\hline
\# of Training QA Pairs & 183,601\\
\# of Question w/ Numbers  & 59,262 (32.3\%)\\	
\# of Answer w/ Numbers  & 73,323 (39.9\%)\\
Avg. Question Length  & 23.0 Chinese characters\\
Avg. Answer Length & 29.6 Chinese characters \\
\end{tabular}
\caption{Dataset statistics.} % Length is measured in Chinese characters.}
\vspace{-10pt}
\label{stat-dataset}
\end{table}

\paragraph{Dataset.}
We collected the question-answer text pairs from Chinese online stock forums.\footnote{\url{http://www.cf8.com.cn/},  \url{http://live.9666.cn/}, \url{http://licaishi.sina.com.cn/}.}
On these forums, users could post their questions about the stocks for free or at very low price. 
%Some stock analysts will answer these questions with relatively short answers. 
%Some representative QA pairs could be found in Table ~\ref{examples-qa}. 
Factoid questions are not encouraged on these sites, as it could be easily found on search engines or financial websites. 
To get a relatively clean dataset, we only keep the QA pairs that contain one single involved stock.
% In other words, the questions containing multiple stocks or no stock are filtered out.
We also delete a QA pair if its answer contains a different stock other than the one mentioned in the question. 
%The stock name or code is replaced with one special token {\em \#STOCK\#} in the question and answer.
Table~\ref{stat-dataset} lists the statistics about the data we built. 
As seen, 32.3\% of questions and 39.9\% of answers contain numbers, which indicate number processing is a critical component of StockQA.

\section{Related Work}

\paragraph{Knowledge Base Question Answering.}
The task of knowledge base question answering has a long history, which refers to constructing answers by querying a knowledge base~\cite{bian2008finding,cong2008finding,berant2013semantic,yao2014freebase,bordes2015large,xu2016question,Yin:2016:IJCAI}. Knowledge base contains facts expressed in various forms, such as factual sentences~\cite{Weston:2015:ICLR,Sukhbaatar:2015:NIPS}, logical form~\cite{berant2013semantic}, and relation triples~\cite{Yin:2016:IJCAI,Miller:2016:EMNLP}.
In this work, we focus on numerical knowledge base, which consists of float numbers in different scales, and thus requires a specific component on number processing. In addition, most previous work aim at retrieving answers -- generally entities, from the knowledge base~\cite{bordes2015large,Weston:2015:ICLR,Sukhbaatar:2015:NIPS,Miller:2016:EMNLP}. In contrast, we generate answers in the form of natural language sentences based on the information from the knowledge base. %, and the entities in the knowledge base cannot be directly used to construct the answer. 

\paragraph{Data to Text Generation.}
In recent years, there has been a growing interest in generating text to describe structured data, such as weather forecasts\cite{goldberg1994,reiter2005choosing,belz2007,angeli2010} and image captioning\cite{vinyals2015show}. One traditional method is using hand-crafted rules\cite{goldberg1994,dale2003coral,reiter2005choosing}. More recently, automatically generating text from series or structured data is widely studied by exploiting neural network based models \cite{vinyals2015show}. We take a further step on this direction by generating interpretations based on structured data as well as text (\ie user questions).

Large-scale numerical data is commonly available in various industries, such as pharmacy, games, finance, and telecommunications.  Interpreting such data is in high demand, while manual interpretation would be prohibitively expensive, both time-wise and financially. 
Several researchers turned to automatic generation of descriptions on numerical data with neural network models~\cite{mei2015talk,murakami2017learning}.
Our work is closely related to stock market comment generation~\cite{murakami2017learning}, which describes the change of only the market stock prices that are generally in the same scale. In contrast, our task is more difficult since (1) the numbers (\eg prices of different stocks) are not in the same scale, which poses difficulty to capturing relation between them; and (2) the numbers in the knowledge base cannot be used to construct the answer with copy or arithmetic operations.

\section{Approach}

\subsection{Architecture}

\begin{figure*}[t]
\centering
\includegraphics[width=0.6\textwidth]{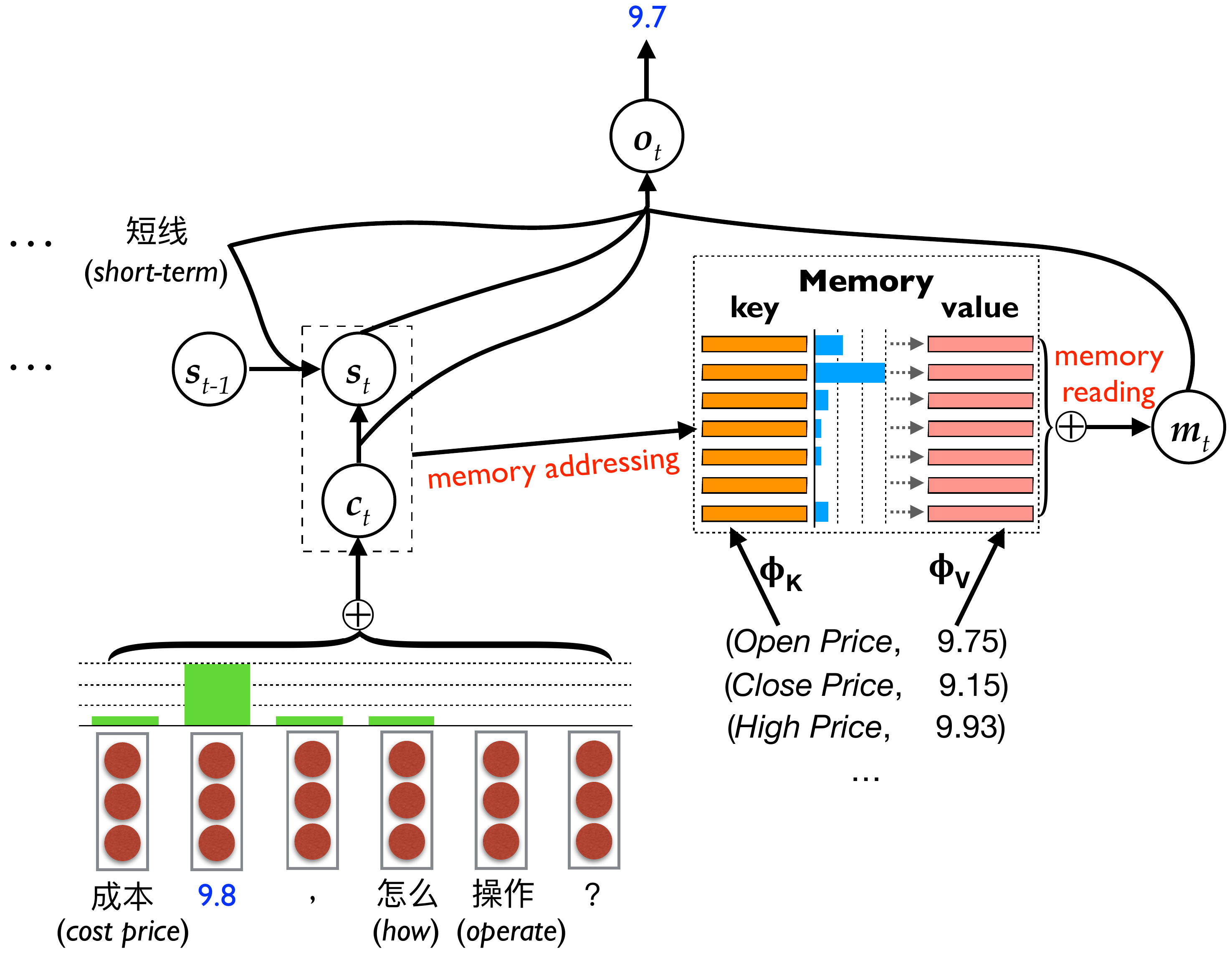}
\caption{Architecture of the proposed approach. % for generating an answer ``短线~9.7~,~站稳~则~持有~({\em Hold it if the price could rise and hold steady above \$9.7 in short-term})''.
At each decoding step, the vector ${\bm c}_t$ that represents the current focus of question and the decoder state ${\bm s}_t$ that summarizes the previously generated words server as a query to retrieve the memory, and a value vector ${\bm m}_t$ that embeds the related stock knowledge is returned for generating the target word.}
\vspace{-5pt}
\label{figure-architecture}
\end{figure*}

As shown in Figure~\ref{figure-architecture}, we employ a memory-augmented encoder-decoder architecture for the StockQA task, which consists of three components:
\begin{itemize}
  \item {\em encoder} that summarizes the question into a sequence of vector representations;
  \item {\em memory} that stores stock knowledge-base as an array of representation pairs;
  \item {\em decoder} that generates the answer word-by-word based on the representations from both the encoder and memory.
\end{itemize}

% \noindent{\em Notation}. to be modified... Our model assigns probabilities to se- quences of words w = w1,...,w|w|, where |w| is the length, and where each word wi is represented by a sequence of characters ci = ci,1, . . . , ci,|ci| of length |ci|.

\vspace{-5pt}
\paragraph{Encoder.}
Suppose that ${\bf x}=x_1, \dots x_j, \dots x_J$ represents a question to answer. The encoder is a bi-directional Recurrent Neural Network (RNN)~\cite{Schuster:1997:TSP} to encode the question ${\bf q}$ into a sequence of hidden states ${\bf H}={\bf h}_1, \dots, {\bf h}_j, \dots {\bf h}_J$, where ${\bf h}_j$ is the annotation of ${\bf x}_j$ from the bi-directional RNN.

%\begin{align}
%{\bf h}_j &= \left[\begin{array}{c}\overleftarrow{\bf h}_j\vspace{3pt}\\\overrightarrow{\bf h}_j\end{array}\right]   &&   \in \mathbb{R}^{2m}\\
%\overrightarrow{\bf h}_j &= f({\bf W}_{h}^1 \overrightarrow{\bf h}_{j-1} + {\bf W}_{q}^1~{\bf e}({q}_j))   && \in \mathbb{R}^{m}\\
%\overleftarrow{\bf h}_j &= f({\bf W}_{h}^2 \overleftarrow{\bf h}_{j+1} + {\bf W}_{q}^2~{\bf e}({q}_j))   && \in \mathbb{R}^{m}
%\end{align}
%where 
%\begin{itemize}
%  \item $f(\cdot)$ is a function to compute the current RNN state given all the related inputs. It can be either a vanilla RNN unit using $\tanh$ function, or a sophisticated gated RNN unit such as GRU~\cite{Cho:2014:EMNLP} or LSTM~\cite{Hochreite:1997}.
%  \item ${\bf e}(q_j) \in \mathbb{R}^{d}$ is an $d$-dimensional embedding of the word $q_j$.
%  \item $\{{\bf W}_{h}^1,{\bf W}_{h}^2\} \in \mathbb{R}^{m \times m}$, $\{{\bf W}_{q}^1,{\bf W}_{q}^2\} \in \mathbb{R}^{m \times d}$ are parameter matrices with $m$ being the numbers of units of encoder RNN state.
%\end{itemize}

\vspace{-5pt}
\paragraph{Memory.}
The memory is essentially a key-value memory network~\cite{Miller:2016:EMNLP}, which is an array of slots in the form of ({\em key}, {\em value}) pairs.
The key layer ${\bf K}=\{{\bf k}_1, \dots, {\bf k}_l, \dots {\bf k}_L\}$ is used for calculating the attention distribution over the memory slots, and the value layer ${\bf V}=\{{\bf v}_1, \dots, {\bf v}_l, \dots {\bf v}_L\}$ is used for encoding stock knowledge representation.
Each key-value pair (${\bf k}_l, {\bf v}_l$) of the stock knowledge corresponds to a distinct stock feature $(k_l, v_l)$, where $k_l$ and $v_l$ are the name and value of the $l$-th stock feature respectively.
In this work, we use word embedding as the key embedding matrix: ${\bf k}_l = e(k_l)$, and we discuss choice of embedding numeric feature values later.

\vspace{-5pt}
\paragraph{Decoder.} 
The decoder is another RNN to generate the answer word by word. 
At each step $t$, the decoder first selects part of the question to focus on:
\begin{equation}
{\bf c}_t = \sum_{j=1}^{J} \alpha_{t,j} {\bf h}_j
% \alpha_{t,j} &= \frac{exp(\tanh({\bf h}_j, {\bf s}_{t-1}))}{\sum_{j'}^{J}{exp(\tanh({\bf h}_j', {\bf s}_{t-1}))}}
\end{equation}
where $\alpha_{t,j}$ is alignment probability computed from a comparison of the previous decoder state ${\bf s}_{t-1}$ and the question annotation ${\bf h}_j$ using a {\em question attention model}~\cite{Bahdanau:2015:ICLR}.
The decoder RNN state ${\bf s}_t$ is updated as
\begin{equation}
{\bf s}_t = f(e(y_{t-1}), {\bf s}_{t-1}, {\bf c}_t)
\label{eqn-hidden-state}
\end{equation}
where $f(\cdot)$ is an activation function, and ${\bf e}(y_{t-1})$ is the word embedding of the previously generated word $y_{t-1}$. Given that ${\bf c}_t$ represents the current focus of the question and ${\bf s}_t$ encodes the partially generated answer, we use both of them to retrieve related records stored in the memory:
%Specifically, the memory attention weight ${\boldsymbol \beta} \in \mathbb{R}^{L}$ are computed as
\begin{align}
{\bf m}_t &= \sum_{l=1}^{L} \beta_{t,l} {\bf v}_l  \\
\beta_{t,l} &= \frac{exp(\tanh({\bf k}_l,  {\bf c}_t, {\bf s}_{t}))}{\sum_{l'}^{L}{exp(\tanh({\bf k}_l',  {\bf c}_t, {\bf s}_{t}))}}
\end{align}
where $\beta_{t,l}$ is alignment probability calculated with another {\em memory attention model}. 
The probability of generating the {\em t}-th word $y_t$ is computed by
\begin{equation}
P(y_t| y_{<t}, {\bf x}) = g(y_{t-1}, {\bf s}_t, {\bf c}_t, {\bf m}_t)
\label{eqn-standard-probability}
\end{equation}
where $g(\cdot)$ first linearly transforms its input and then applies a softmax function.

%The parameters associated with the proposed architecture are trained to maximize the {\em likelihood} of a set of training examples $\{\left[{\bf Q}^i, {\bf A}^i, {\bf S}^i\right]\}_{i=1}^{I}$:
%\begin{eqnarray}
%\mathcal{L}(\theta) &=& \argmax_{\theta}\sum_{i=1}^{I}\log P({\bf A}^i|{\bf Q}^i, {\bf S}^i; \theta) \nonumber \\
%                    &=& \argmax_{\theta}\sum_{i=1}^{I}\sum_{t=1}^{T}\log P(y_t|y_{<t}, {\bf Q}^i, {\bf S}^i; \theta)
%\label{eqn-standard-training}
%\end{eqnarray}

\subsection{Number Encoding and Decoding}

StockQA is an open-vocabulary problem, since the vocabularies of numbers keep growing as the number of stocks grows and stock price changes over time.
Nearly all neural generative models have used quite restricted vocabularies, crudely treating all other words the same with an ``$<$UNK$>$'' symbol, called Out-of-Vocabulary (OOV) words. 
In this task, numbers in both question-answer pair and stock knowledge base are low-frequent or even unseen in the training corpus and thus typically treated as OOV words.
The root of the OOV problem lies in the complete inability of word-level neural models to sensibly assign nonzero probability to previously unseen words (\ie numbers in this task).

Inspired by recent successes of smaller granularity on alleviating the OOV problem~\cite{Luong:2016:ACL,Sennrich:2016:ACL,Lee:2017:TACL}, we decompose numbers into characters and construct the meaning of numbers (in both question and knowledge base) compositionally from characters, as well as generate numbers (in answer) character by character. 
In this work, we propose two types of word-character models, which differ at how the character model is integrated into the word-level encoder-decoder framework.

\vspace{-5pt}
\paragraph{Sequential Word-Character Model.}
A simple way is to decompose numbers into smaller granularities, such as characters~\cite{Bahdanau:2016:ICASSP,Lee:2017:TACL}
%~\cite{Bahdanau:2016:ICASSP,CostaJussa:2016:ACL,Chung:2016:ACL,Lee:2017:TACL} 
or subwords~\cite{Luong:2013:CoNLL,Sennrich:2016:ACL}.
%~\cite{Mikolov:2012:Preprint,Luong:2013:CoNLL,Sennrich:2016:ACL}.
% Figure~\ref{figure-number-generation}(a) shows an example at character level. 
We split numbers at both sides into characters, and directly feed the characters to the encoder and decoder RNNs. Ideally, the encoder RNN is able to construct the meaning of numbers compositionally from characters, and the decoder RNN is able to generate numbers character by character. We use an independent character RNN to encode numbers in knowledge base and use the last hidden states as their representations.

A potential problem of this strategy is that encoder and decoder states need to embed information for (1) question answering and (2) number composition and decomposition. 
%implicitly serves as several functions at the same time:
%\begin{itemize}
%    \item [1.] They embed necessary information for understanding questions, retrieving information from memory, and generating answers.
%    \item [2.] They model the composition of character sequences and the decomposition of numbers.
% \end{itemize}
We hypothesize that such overloaded use of encoder and decoder states makes training the model difficult.
Inspired by recent work~\cite{Rocktaschel:2017:ICLR}, we employ a hybrid scheme~\cite{Luong:2016:ACL} to explicitly separate the two types of functions. %, which has proven successful in machine translation.

\begin{figure}[t]
\centering
      \includegraphics[width=0.3\textwidth]{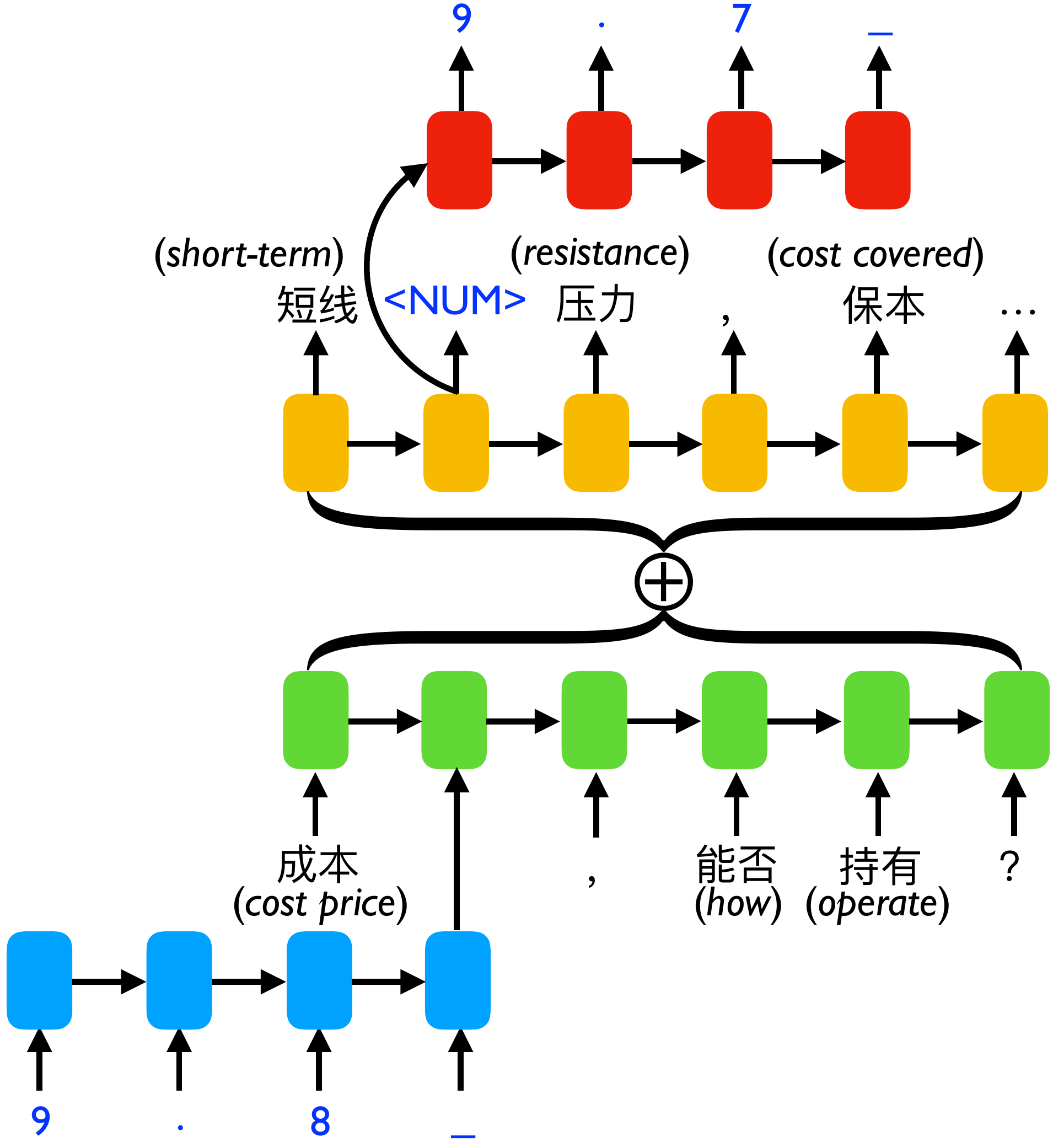}
\caption{A hybrid word-character models for number generation, in which numbers are highlighted in {\color{blue} blue} color. For simplicity, we skip the memory component. }
\label{figure-number-generation}
\vspace{-5pt}
\end{figure}

\vspace{-5pt}
\paragraph{Hybrid Word-Character Model.}
The hybrid model introduces two additional networks to consult character components whenever the model encounters a number, as shown in Figure~\ref{figure-number-generation}.
Specifically, the character-level encoder builds representations for numbers on-the-fly from character units, while the character-level decoder generates numbers character by character based on the corresponding word-level decoder state.
For example, the character encoder updates its state over `9', `.', `8', and `\_' (the boundary tag). The final hidden state will be used as the representation for the current number, which is fed to the word-level encoder RNN.

To enable character-level generation for numbers, we replace all numbers % in answers 
with a special symbol ``$<$NUM$>$''. When the word-level model generates ``$<$NUM$>$'', we use another separate RNN to generate the character sequence given the current word-level state as well as the weighted memory value vector.
We expect the value vector retrieved from the memory can guide the character-level decoder to generate reasonable numbers.\footnote{In our preliminary experiments, the generated numbers are not reasonable without a signal from the memory value.}
We train the hybrid model by consulting this character-level decoder to recover the correct surface form of numbers. Accordingly, the training objective is $\mathcal{L} = \mathcal{L}_w + \lambda\mathcal{L}_c$, where $\mathcal{L}_w$ and $\mathcal{L}_c$ are the likelihoods of the word-level and character-level models, respectively.

\section{Experiment}

\subsection{Setup}

%\begin{table}[t]
%\centering
%\begin{tabular}{c|rr c rr}
%\multirow{2}{*}{\bf Segmentation}	&	\multicolumn{2}{c}{\bf Question}	&	&	\multicolumn{2}{c}{\bf Answer}\\
%\cline{2-3}\cline{5-6}
%        &   \# tokens    &   \# types     &&  \# tokens    &    \# types\\
%\hline
%Word    &   2.02M   &   20.1K   &&  2.49M   &   25.6K\\
%BPE     &   1.49M   &   5.0K    &&  2.10M   &   5.0K\\
%\end{tabular}
%\caption{Corpora statistics (exclude numbers) of different segmentation strategies.} %K stands for thousands and M for millions.}
%\label{table-statistics}
%\end{table}

\noindent {\bf Data Processing.}
Since the QA pairs are in Chinese, we need to segment the sentences into words. We used two strategies: (1) standard word segmentation which is trained on Chinese Treebank -- a domain highly different from the stock data; and (2) Byte Pair Encoding (BPE)~\cite{Sennrich:2016:ACL}: %-- a simple data compression technique, which has been successfully adapted for word segmentation by iteratively replacing the most frequent pair of characters in a sequence with a single, unused symbol. 
we split the question and answer sentences by numbers and punctions,
% (\eg \begin{CJK*}{UTF8}{gbsn}``，'', ``。'', ``；''\end{CJK*}, etc.), 
and subsequently apply the BPE algorithm to these fragments. The intuition behind is that there exist a large number of terminologies and collocations in the answers, which can not be captured by the standard word segmenter. We expect that the BPE algorithm can automatically mine them based on their frequent occurrences.
The word segmentation splits the corpus into 2.02M question tokens and 2.49M answer tokens,
% while the numbers of BPE are 1.49M and 2.10M. % splits into 1.49M question tokens and 2.10M answer tokens.
while BPE splits into 1.49M question tokens and 2.10M answer tokens.
We limited the source and target vocabulary size to 5000 for both strategies, which cover 98.9\% and 99.2\% words for word segmentation, and all tokens for BPE.
%Table~\ref{table-statistics} lists the corpus statistics of the two segmentation strategies. 
%For the standard word segmentation, the vocabularies cover 98.9\% and 99.2\% words in question and answers respectively. 
%For the BPE algorithm, the vocabularies, which consist of phrases, words and subwords, cover all the tokens.

\noindent {\bf Evaluation.}
We randomly extracted 1000 sentence pairs as the validation set, and another 500 sentence pairs as the final test set. We used an automatic metric -- 2-gram BLEU~\cite{Papineni:2002}, to select the best trained model on the validation set, and used human evaluation on the test set to report convincing results. 
Specifically, we asked two annotators to label the answers using the pre-defined metrics.

\noindent {\bf Model.}
We used a bidirectional LSTM layer for encoding and one LSTM layer for decoding. 
The encoder LSTM layer consists of 128 hidden units, and the decoder LSTM layer consists of 256 hidden units. 
The dimensionality of word embeddings, memory key layer and value layer is 128.
When character model is applicable, the dimensionality of character embeddings, encoder layer and decoder layer is 32, 64, and 128, respectively;
%character embedding is 32, the character-level encoder LSTM layer consists of 64 hidden units, the character-level decoder LSTM layer consists of 128 hidden units, and 
the interpolation weight $\lambda=1$. 
%Given that the memory key is the word embedding of feature name, the dimensionality of key layer is 128. The dimensionality of memory value layer is $32L$ ($L$ is the number of features) if bit representation is applied, otherwise is 128.
We used Adam~\cite{Kingma:2014:arXiv} with a learning rate of $0.01$ and the batch size is 128.

\subsection{Generative Model}

\begin{table*}[t]
\centering
\renewcommand\arraystretch{1.1}
\begin{tabular}{c|c|c||ccccc|c}
\multirow{2}{*}{\bf \#} &   \multirow{2}{*}{\bf Segmentation}    &   \multirow{2}{*}{\bf \specialcell{Number\\Enc-Dec}}   &    \multicolumn{5}{c|}{\bf Score Distribution}   &   \multirow{2}{*}{\bf \specialcell{Total\\Score}}\\
\cline{4-8}
            &   &   &   0   &   0.25   &   0.5   &   0.75   &   1.0   &  \\
\hline
1   &   Word    &   Sequential     &  7.3\%   &   19.5\%  &   33.5\%  &   21.6\%  &   18.1\%  &   0.56\\
\hline
2   &   \multirow{2}{*}{BPE}	&	Sequential    &   5.6\% &   17.4\%  &   37.2\%  &   22.4\%  &   17.4\%  &   0.57\\
3   &        &   Hybrid  &   4.8\%   &   15.4\%  &   28.6\%  &   29.6\%  &   21.6\%  &   0.62\\
% \hline
% 4   &   \multicolumn{2}{c||}{Retrieved Answers} &  4.4\%    &   20.2\%   &   9.6\%   &   54.0\%  &   11.8 \%  &  0.62 \\
% \hline
% 5   &   \multicolumn{2}{c||}{Multi-Source} &   1.2 \%    &   19.6\%   &   7.4\%   &   56.0\%  &   15.8\%  &   0.66\\
\hline
\hline
4   &   \multicolumn{2}{c||}{Human Generated Answers} &   1.0\%    &   7.8\%   &   7.0\%   &   24.2\%  &   60.0\%  &   0.84\\
\end{tabular}
\caption{Manual evaluation of answers generated by the proposed generative models and humans. A score of ``0'' denotes low-quality answers that are not relevant to the questions, while ``1'' denotes high-quality answers perform well in all the metrics.}
%\vspace{-10pt}
\label{table-results}
\end{table*}

\begin{figure*}[t]
\centering
\vspace{-10pt}
\subfloat[\small Question attention matrix.]{
    \includegraphics[width=0.23\textwidth]{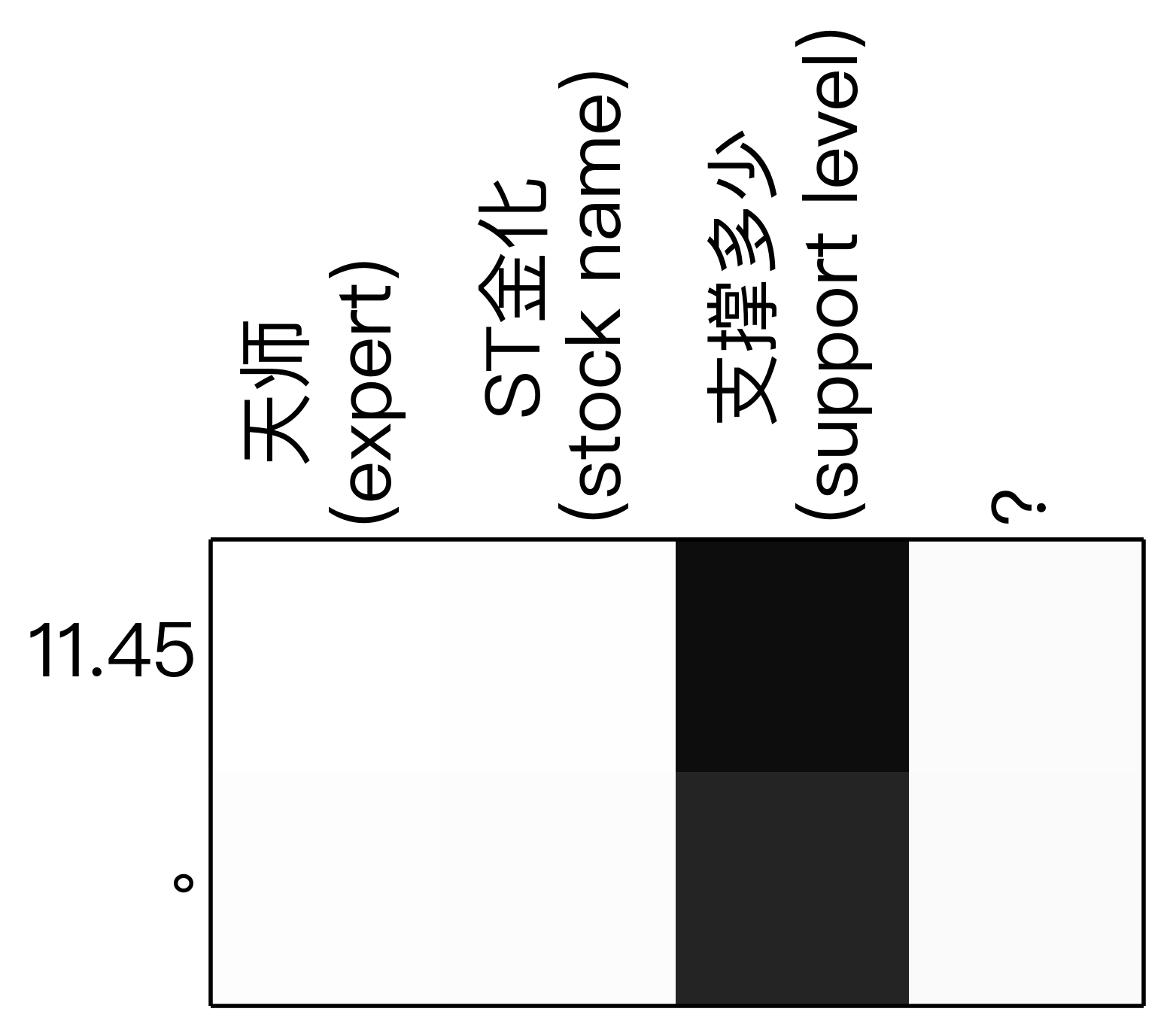}} \hfill
\subfloat[\small Memory attention matrix.]{
    \includegraphics[width=0.7\textwidth]{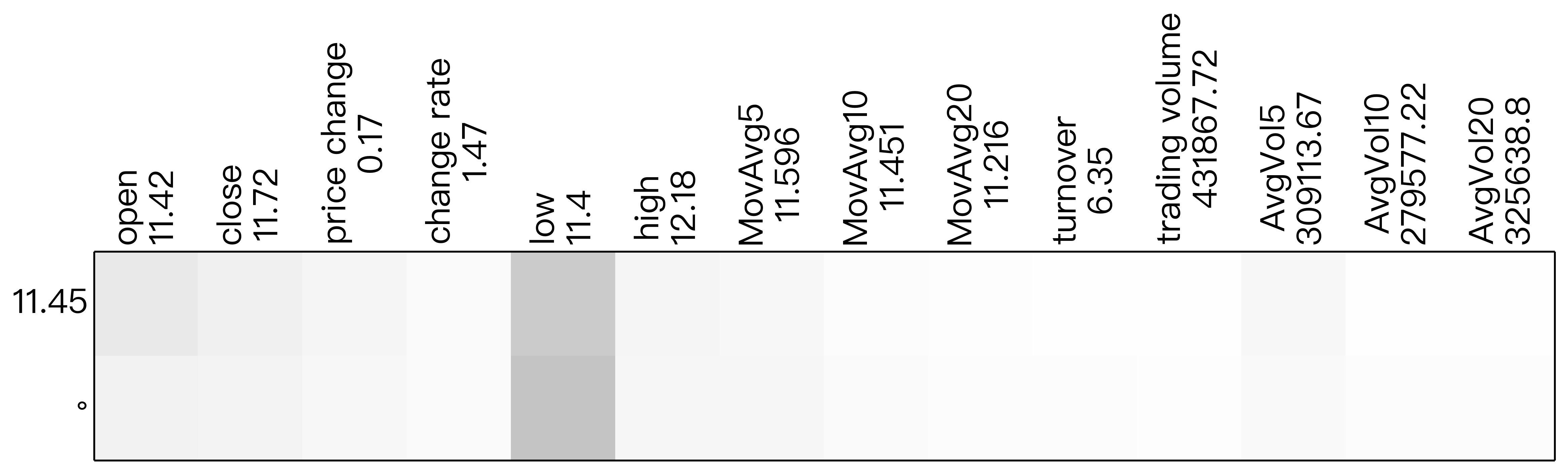}}
\caption{Visualization of attention matrices. $y$-axis is the generated answer (\ie 
``11.45 .'') and $x$-axes are (a) question ({\em ``Expert, what is the support level of \#STOCK\#?''}) and (b) numeric stock knowledge.} %``MovAvg$k$'' denotes moving average over $k$ days, and ``AvgVol$k$'' denotes average trading volume of $k$ days.}
\vspace{-5pt}
\label{figure-attention}
\end{figure*}

\noindent{\bf Manual Evaluation} 
Table~\ref{table-results} shows the manual evaluation scores on the test set.
% \noindent{\bf Segmentation Strategies} (Rows 1 vs. 2) 
We first compare the segmentation strategies with sequential word-character model for number processing (Rows 1 vs. 2). As seen, the BPE algorithm outperforms the standard word segmenter by dividing the sentence into larger granularities, which  eases the understanding of relatively ``shorter'' questions and thereby reduces the number of irrelevant and imprecise answers (\ie $Score \leq 0.25$). %In the following experiments, we use BPE as the default segmentation strategy.

%\noindent{\bf Numerical Feature Representation Strategies} (Rows 2 vs. 3, 4 vs. 5)
%The two types of representations achieve comparable performances when using sequential word-character model for number generation. In the context of hybrid generation model, the continuous representation outperforms its bit counterpart, since the former is more compatible with hybrid model as they use the same character-level RNN to embed float numbers that occur in both questions and stock knowledge base.

% \noindent{\bf Number Processing Strategies} (Rows 2 vs. 3)
Clearly, the hybrid word-character model significantly outperforms its sequential counterpart (Rows 2 vs. 3), especially on the portions of good answers (\ie $Score \geq 0.75$). We attribute this to the additional character-level neural networks, which separates the processing of numbers from encoder and decoder states. We expect that this explicit design would lead to a better generation of numbers, which we will validate later through experiments.

% \noindent{\bf Answers Generated by Humans} (Row 6)
To make sure that the evaluation metrics are reasonable and convincing, we also ask the annotators to evaluate the answers generated by humans. As seen, 84.2\% of human answers are labelled as high-quality (\ie $Score\geq 0.75$), proving that the evaluation metrics are reasonable. On the other hand, there are still 15.8\% of the human answers is poor in at least one metric (\ie $Score\leq0.5$), which poses difficulty to the model training by being noisy instances. The performance gap between answers generated from the proposed model and humans (0.62 vs. 0.84), indicates large room for further improvement.

\begin{table}[t]
\centering
\begin{tabular}{c|c||cc|cc}
\multirow{2}{*}{\bf Seg.}    &   \multirow{2}{*}{\bf \specialcell{Number\\Enc-Dec}}   &    \multicolumn{2}{c|}{\bf w/ Number}   &   \multicolumn{2}{c}{\bf w/o Number}\\
\cline{3-6}
                        &   &   \em Perc.  &  \em   Score   & \em Perc. & \em  Score  \\
\hline
Word                    &   Seq.     &  10.4\%  &   0.66    &   89.6\%  &   0.55\\
\hline
\multirow{2}{*}{BPE}	&   Seq.     &  11.2\%  &   0.70    &   88.8\%  &   0.56\\
        				&   Hyb.     &  38.6\%  &   0.73    &   61.4\%  &   0.55\\
\hline
\hline
\multicolumn{2}{c||}{Human} &   45.0   &   0.93    &   55.0  &   0.77\\
\end{tabular}
\caption{Performance on number generation. We divide the generated answers into two parts by whether they contain numbers (``w/ Number'') or not (``w/o Number'').}
\label{table-number-generation}
\end{table}

\vspace{-5pt}
\paragraph{Contribution Analysis}
In this experiment, we investigate whether the improvement of hybrid model comes from better performance on number generation. To this end, we divide the generated answers into two parts by whether they refer to number generation. As shown in Table~\ref{table-number-generation}, the proposed models achieve similar performances on generating answers that do not contain numbers, although there are still considerable differences on the portions the proposed variants account for. Concerning number generation, the hybrid model outperforms its sequential counterpart in terms of both quantity and quality, which validates our belief. 
% Table~\ref{table-number-cases} shows some examples, in which the hybrid model is more prone to generate answers containing numbers.

\iffalse
\begin{figure}[h]
\centering
\subfloat[\footnotesize Question attention.]{
    \includegraphics[width=0.18\textwidth]{figures/attention-matrix-question.png}} \\
\subfloat[\footnotesize Memory attention matrix.]{
    \includegraphics[width=0.47\textwidth]{figures/attention-matrix-memory.png}}
\caption{Visualization of attention matrices. $y$-axis is the generated answer (\ie 
``11.45 .'') and $x$-axes are (a) question ({\em ``Expert, what is the support level of \#STOCK\#?''}) and (b) numeric stock knowledge.} %``MovAvg$k$'' denotes moving average over $k$ days, and ``AvgVol$k$'' denotes average trading volume of $k$ days.}
\label{figure-attention}
\end{figure}
\fi

Figure~\ref{figure-attention} shows an example on number generation. Given a question on support level that generally refers to the low price of a stock, the word-level decoder first selects the corresponding question context with a question attention, and subsequently retrieve the information about low price from the memory. Based on both context vectors, the character-level decoder is able to generate a float number ``11.45'', which is close to the low price ``11.4'' as expected.

% In the following experiments, we do the analysis on the representative model ``BPE + Continuous Representation + Hybrid Model''.

\begin{table}[t]
\centering
\begin{tabular}{c|rc}
{\bf Error Type}    &   {\bf Portion}     &   {\bf Score}\\
\hline
Poor in at least one metric   &   48.8\%   &   0.37\\
\hline
Poor in Relevance           &   4.8\%   &   0\\
Not Fluent                  &   1.2\%   &   0.21\\
Has Factual Errors          &   8.6\%   &   0.14\\
Poor in Informativeness     &   44.8\%  &   0.38\\
\end{tabular}
\caption{Portions and scores of generated answers with different error types. Portion numbers are absolute percentage over all instances in the test set.}
\vspace{-8pt}
\label{table-error-types}
\end{table}

\paragraph{Error Analysis}
From Table~\ref{table-results}, we found that 48.8\% of answers generated by our best model (Row 3) still perform poor in at least one of the four metrics (\ie $Scores\leq0.5$). Table~\ref{table-error-types} lists the statistics on detailed error types. Note that the sum of portions on each error type is not be equal to 48.8\%, since an answer may have more than one type of errors (\eg poor in both relevance and informativeness).
The majority of error answers comes from poor informativeness, which is mainly caused by producing generic answers -- a common problem in dialogue and question answering tasks with neural network models~\cite{Sordoni:2015:ICIKM,Vinyals:2015:DLW,Serban:2016:AAAI,Li:2016:NAACLa}.

\subsection{Hybrid Retrieval-Generation Model}

We expect that the generic answer problem can be alleviated by imitating similar answers retrieved from the training corpus, which are high quality samples: they are natural sentences that exhibit no bias towards shortness or vagueness. An straightforward implementation of such idea is a hybrid retrieval-generation model, which uses an additional set of encoder and attention model to encode and select part of the retrieved answer for generating each target word.

Formally, let ${\bf \hat{y}} = \{\hat{y}_1, \dots, \hat{y}_I\}$ be the retrieved answer, the probability of generating the $t-$th target word $y_t$ is rewritten as
\begin{equation}
P(y_t| y_{<t}, {\bf x}, {\bf \hat{y}}) = g(y_{t-1}, {\bf s}_t, {\bf c}_t, {\bf m}_t,  {\bf r}_t)
\label{eqn-retrieval-probability}
\end{equation}
where ${\bf r}_t$ is the context vector calculated by an separate attention model over the representations of retrieved answer.

We retrieved answers from the training answers based on the similarities of both the question and stock knowledge. We use Jaccard distance to compute the similarity between two question sentences. 
For stock similarity, we first compute the tendency of a stock by the ratios of open prices out of MovAvg5, MovAvg10, and MovAvg20. The ratios of current stock knowledge and the retrieved stock are then used to compute the similarity.

% \usepackage{multirow}
\begin{table}[t]
\centering
\begin{tabular}{c|cc|cc|c}
\multirow{2}{*}{\bf Model} & \multicolumn{2}{c|}{\bf w/ Number} & \multicolumn{2}{c|}{\bf w/o Number}    &   \multirow{2}{*}{\bf All}\\ 
\cline{2-5} 
                       &    Perc.           & Score           &     Perc.           & Score           \\ \hline
Generative             &   38.6\%   &   0.73    &   61.4\%    &   0.55 &   0.62    \\ %\hline
Retrieval              &   24.6\%   &   0.53    &   75.4\%    &   0.65    &   0.62    \\ %\hline
\hline
Hybrid           &   28.4\%   &   0.65    &   71.6\%    &   0.67    &   0.66    \\ %\hline
% Human                  &   45.0\%   &   0.93    &   55.0    &   0.77       \\ \hline
\end{tabular}
\caption{Performance of hybrid retrieval-generation model. ``Generative'' denotes the best generative model in the above section.}
\label{table-number-generation}
\end{table}

\paragraph{Manual Evaluation.}
Table~\ref{table-number-generation} lists the manual evaluation scores of the hybrid retrieval-generation model. As seen, the retrieval answers share the same overall score with the generation model. However, the retrieval model performs better at the answers without numbers, which is opposite to the generative model. This is intuitive, since the numbers in the retrieved answers often conflicts with the values in the stock knowledge, thus has factual error. 
As expected, the hybrid retrieval-generation model combines advantages of both models, and improves the overall performance.

\begin{table}[t]
\centering
\begin{tabular}{c||cccc}
\multirow{2}{*}{\bf Model}   &    \multicolumn{4}{c}{\bf Diversity at $n$-gram}  \\
\cline{2-5}
&   1   &   2   &   3   &   4\\
\hline
Generative                    &   0.18  &   0.27  & 0.31  & 0.33  \\
Retrieval                     &   0.28  &   0.77  & 0.93  & 0.96  \\
Hybrid                        &   0.25  &   0.63  & 0.83  & 0.91  \\
\hline
Human                         &   0.30  &   0.82  & 0.97  & 0.99  \\
\end{tabular}
\caption{Evaluation of diversity at different granularities, ranging from 1-gram to 4-gram. Higher number denotes higher diversity.}
\label{table:diversity}
\end{table}

\paragraph{Diversity Evaluation.}
We analyze the diversity of our model comparing to the three baselines quantitatively and qualitatively. If the diversity for one model is better, the n-grams of the generated sentences will be significantly larger than other models. Starting from this intuition, we count the number of {\em unique} n-grams appears in each answer and normalize it with the total number of n-grams. The result is shown in Table \ref{table:diversity}. As the retrieved answers are searched from the training data, the diversity of the retrieved answers are significantly better than other models. The generation model tends to generate generic responses, which reconfirm our previous findings. By utilizing retrieved answers, the diversity of the hybrid model increases significantly, which shows the benefits of external information source.

\begin{table}[t]
\centering
\begin{tabular}{c||cccc||c}
\multirow{2}{*}{\bf Model}   &    \multicolumn{4}{c||}{\bf Number of $n$-gram}    &   \multirow{2}{*}{\bf Manual}   \\
\cline{2-5}
&   1   &   2   &   3   &    4 \\
\hline
Generative                    &   389   &  466  & 454   &  418  &   1.972\\
Retrieval                     &   1420  & 3430  & 3700  &  3410 &   1.998\\
Hybrid                        &   1302  & 3006  & 3597  &  3560 &   2.114\\
\hline
Human                         &   1634  & 4057  & 4424  & 4152  &   2.532\\

\end{tabular}
\caption{Evaluation of informativeness at different granularities. Higher number denotes higher informativeness. ``Manual'' denotes manual evaluation score.}
\label{table:informativeness}
\end{table}

\paragraph{Informativeness Evaluation.}
We analyze the informativeness of different models based on the size of the unique $n$ grams, as listed in Table \ref{table:informativeness}. We find that the generative model tends to produce short answers, and the hybrid model greatly alleviates this problem by considering information of retrieved answers.

% by utilizing retrievals, the proposed model tends to generate larger size of $n$ grams than the generation model by a large margin. This demonstrates that the generated answers are more informative than the baseline model.

We also list the results of manual evaluation on informativeness, which measures whether the answer provides useful information to humans. The three ratings for informativeness (``poor'', ``normal'', ``good'') corresponds to the scores (1, 2, 3). As seen, although the retrieval model offers long answers in natural sentences, the embedded information may not be always useful for humans. The hybrid model can combine advantages of both models, and improve the informativeness of the generated answers.

\section{Conclusion}

We present a pioneering work on the generative stock related question answering (StockQA) task. One key challenge of this task lies in understanding and generating numbers in different scales, which is the focus of this work. 
We have exploited several mechanisms to learn to compose and decompose numbers. 
Experimental results show that a hybrid word-character model can produce reasonable answers with separate components to explicitly process numbers. 

Although the proposed generative model is able to generate acceptable answers with reasonable numbers, it still suffers from {\em generic answer} problem. 
Our study shows that a hybrid retrieval-generation model can greatly alleviates this problem, and significantly improves the diversity of informativeness of the generated answers.

\bibliographystyle{aaai}
\bibliography{all} 
\end{document}